\documentclass[final,2p,times,twocolumn]{elsarticle}

\usepackage{amssymb}
\usepackage{xcolor}
\usepackage{amsmath}
\usepackage{booktabs}
\usepackage{multirow}
\usepackage{algorithm}
\usepackage{algorithmic}
\usepackage{colortbl}
\usepackage{xcolor}
\usepackage[table]{xcolor}
\newcommand{\com}[1]{\quad{\scriptsize\textcolor{gray}{#1}}}
\usepackage{hyperref}
\usepackage{booktabs,siunitx,makecell}
\usepackage{booktabs,siunitx,makecell}
\begin{document}

\begin{frontmatter}

\title{J-RAS: Mutual Adaptation for Medical Image Segmentation via Contrastive Retrieval-Augmented Joint Optimization}

\author[a]{Salma J. Ahmed\corref{cor1}}
\ead{ahme3460@mylaurier.ca} 

\author[a]{Emad A. Mohammed}
\author[a]{Azam Asilian Bidgoli} 

\cortext[cor1]{Corresponding author.} 

\affiliation[a]{organization={Department of Physics and Computer Science, Wilfrid Laurier University},
            city={Waterloo},
            postcode={N2L 3C5}, 
            state={ON},
            country={Canada}}

\begin{abstract}
Manual medical image segmentation by clinicians, though accurate, is time-consuming and variable across experts, while AI-based models automate this process but often falter under limited data and domain shifts. Inspired by how humans learn new tasks through guidance such as how children draw more accurately when shown reference images, we propose Joint Retrieval-Augmented Segmentation (J-RAS), a framework that enables segmentation networks to learn with guidance. J-RAS jointly optimizes a segmentation model and a retrieval model through alternating contrastive and supervised learning, allowing the retrieval network to discover contextually relevant image–mask pairs that refine the segmentation model’s anatomical reasoning. Unlike conventional retrieval-based augmentation that passively provides similar samples, J-RAS establishes a mutual adaptation and optimization loop where the retrieval model learns to emphasize segmentation-relevant cues, while the segmentation model leverages retrieved examples to improve boundary delineation, robustness to rare cases and and cross-dataset generalization. Evaluations on two public benchmarks, ACDC and M\&Ms, across multiple backbones (U-Net, TransUNet, SAM, and SegFormer) demonstrate the generality and effectiveness of J-RAS. For instance, on ACDC, SegFormer improves from a mean Dice of 0.8708 $\pm$ 0.042 and HD of 1.8130 $\pm$ 2.49 to 0.9115 $\pm$ 0.031 and 1.1489 $\pm$ 0.30. These results highlight how retrieval-guided contrastive optimization bridges human-like guidance and machine-learned precision in medical image segmentation.
\end{abstract}


\end{frontmatter}

\section{Introduction}
\vspace{-5pt}
\label{Intro}
When learning a new skill or task, guidance often plays a pivotal role in improving performance, much like how children learn to draw or color by being shown examples or receiving instruction \cite{sashkin1977social,vygotsky1978mind,honomichl2012role}. For instance, when drawing a bird for the first time, having a reference image leads to more accurate and confident results. Similarly, Artificial Intelligence (AI) systems \cite{fetzer1990artificial} are designed to emulate human cognitive behavior \cite{turing2009computing,russell2016artificial}, learning from experience to perform tasks such as perception, decision-making, and reasoning. In computer vision \cite{stockman2001computer}, the AI models mimic human visual understanding \cite{lecun2015deep}, interpreting complex medical images to segment organs and lesions with precision.

Manual segmentation by healthcare professionals relies heavily on visual perception and prior knowledge to segment organs or pathological regions, drawing precise boundaries to support diagnosis and treatment planning. However, this process is time-consuming, costly, and subject to inter-observer variability \cite{sudre2021image,wang2021annotation,gao2024effective}. Deep learning models, particularly convolutional neural networks (CNNs) \cite{o2015introduction} and vision transformers (ViTs) \cite{dosovitskiy2020image} have improved automation \cite{liu2019recent,litjens2017survey}, yet they remain constrained by dependence on large annotated datasets, sensitivity to domain shifts across scanners or populations \cite{ghafoorian2017transfer}, and lack of prior anatomical knowledge.
\begin{figure*}[ht]
    \centering
    
    \includegraphics[width=\textwidth]{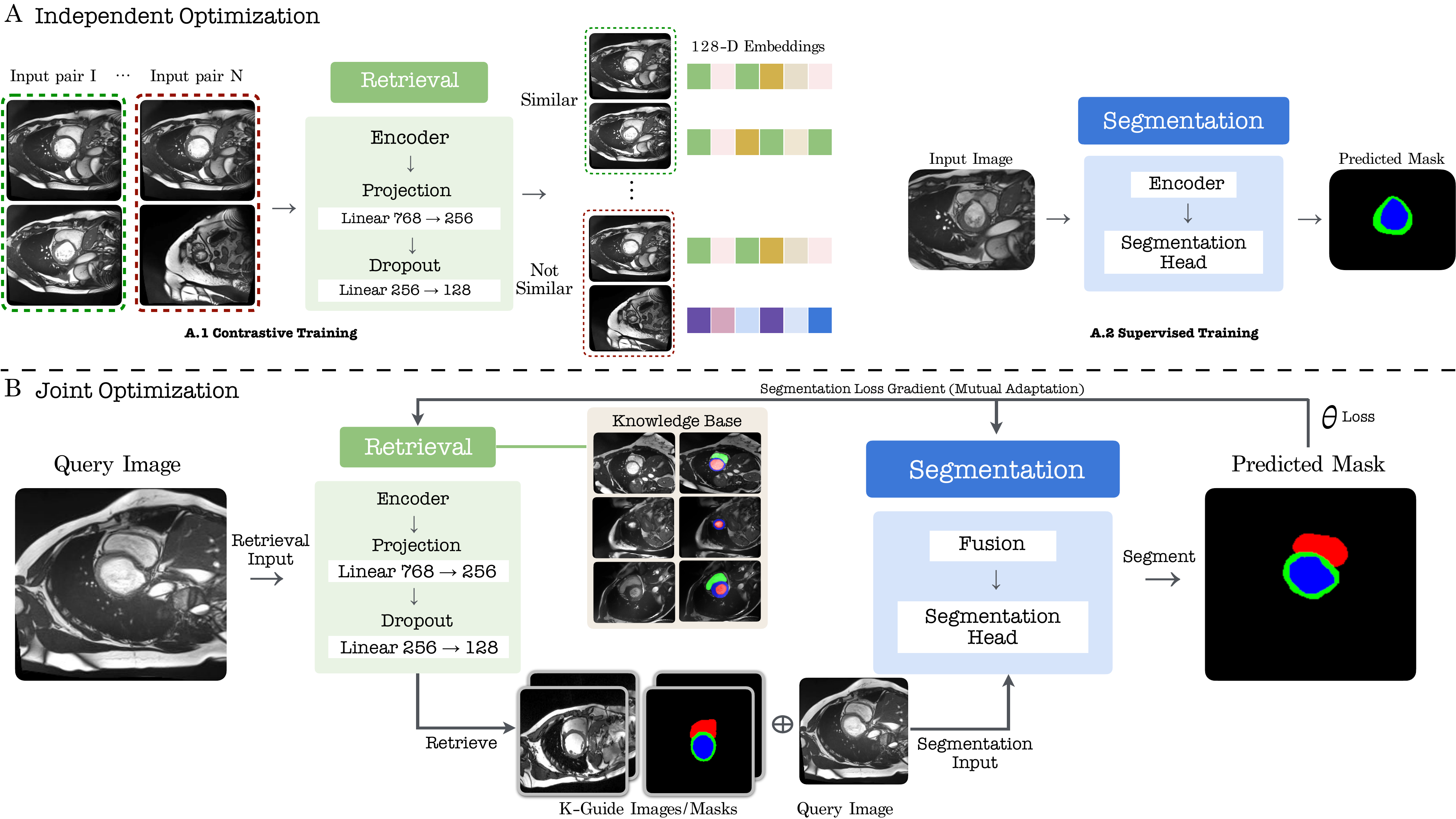}
    \caption{Overview of the proposed Joint Retrieval-Augmented Segmentation (J-RAS) method. \textbf{\textit{Stage A:}} Independent optimization phase. The retrieval model (\textbf{A.1}) is first fine-tuned to learn discriminative image embeddings, while the segmentation model (\textbf{A.2}) is fine-tuned to predict anatomical masks.  \textbf{\textit{Stage B:}} Joint optimization phase. Given a query image, the top-$K$ most similar guide images and their corresponding masks are retrieved from the knowledge base. The query image and retrieved guides are fused and passed to the segmentation model to predict the final mask. The segmentation loss computed on the predicted mask is then used to update both the segmentation and retrieval models simultaneously.}
   \vspace{-4mm}
    \label{fig:pipeline} 
 
\end{figure*}

In clinical settings, radiologists often consult similar past cases to inform decisions \cite{chen2023investigating}. This reasoning parallels medical image retrieval \cite{qayyum2017medical}, which surfaces anatomically or pathologically related scans from large databases. Prior work \cite{zhao2025retrieval} attempted to combine retrieval with segmentation, but it treated the retrieval model as a fixed feature extractor. If the retrieved samples fail to assist in segmentation, the system cannot adapt or learn from that failure. Consequently, the guidance remains static rather than interactive, limiting its potential to improve downstream performance.

To overcome this limitation, we propose \textbf{J}oint \textbf{R}etrieval-\textbf{A}ugmented \textbf{S}egmentation (J-RAS), a unified framework that enables segmentation networks to learn with guidance. J-RAS jointly optimizes both the retrieval and segmentation models through alternating contrastive and supervised learning, forming a mutual adaptation loop, the retrieval model learns to prioritize segmentation-relevant cues, while the segmentation model dynamically incorporates retrieved image–mask pairs to refine boundary precision, anatomical reasoning, and cross-dataset generalization.

We validate J-RAS on \textbf{four} backbones, namely U-Net \cite{ronneberger2015u}, TransUNet \cite{chen2021transunet}, SegFormer \cite{xie2021segformer}, and SAM \cite{kirillov2023segment} and \textbf{two} benchmark datasets, ACDC \cite{bernard2018deep} and M\&Ms \cite{campello2021multi}. Evaluation covers Dice coefficient, Hausdorff Distance (HD), and qualitative visualizations demonstrating improvements in boundary delineation, mask consistency, and retrieval interpretability.

Additionally, we conduct extensive experiments to evaluate the robustness and adaptability of J-RAS. Specifically, we perform ablation studies on fusion strategies for integrating guidance (\autoref{sec:fusion}), compare dynamic Top-K retrieval with fixed sampling (\autoref{sec:DynamicK}) to test its effect on relevance, and assess robustness by introducing noise into the guidance (\autoref{sec:Noisee}). We further examine cross-dataset generalization, where retrieval and segmentation are trained on different datasets (\autoref{sec:cross_in_gen}), and include a detailed per-case analysis to evaluate behavior across patients (\autoref{sec:case_analysis}). Our contributions are summarized as follows:

\begin{enumerate}
    
    \item We propose \textbf{J-RAS}, a retrieval-guided segmentation framework that, for the first time, jointly optimizes retrieval and segmentation through alternating contrastive and supervised learning.
 
    \item J-RAS transforms retrieval from a passive feature provider into an \textbf{adaptive collaborator}, enabling mutual feedback that enhances segmentation-relevant representations.

    \item We demonstrate \textbf{architecture-agnostic performance} and \textbf{cross-dataset generalization} (\autoref{sec:main_res}, \autoref{sec:cross_in_gen}), showing that J-RAS can retrieve from one dataset while segmenting another.

    \item Extensive \textbf{quantitative, qualitative, and ablation studies} demonstrate that J-RAS consistently outperforms baseline models in boundary precision, robustness to noise, and contextual interpretability.
    
\end{enumerate}

\section{Related Work}
Image segmentation is a well-known and extensively studied problem in the medical artificial intelligence domain, with numerous methods proposed over the years. 

\vspace{5pt}
\textbf{Classical Approaches.} often rely on CNN architectures \cite{o2015introduction}, such as U-Net \cite{ronneberger2015u} and its variants. More recently, transformer-based models \cite{vaswani2017attention}, including SegFormer \cite{xie2021segformer} and Swin-Unet \cite{cao2022swin}, have been introduced to capture richer global context. Hybrid architectures, such as TransUNet \cite{chen2021transunet}, combine the strengths of U-Net and transformers to leverage both local detail and long-range dependencies. However, these methods can still face challenges in handling limited data availability, large variations between patients, and rare pathological cases.

\vspace{5pt}
\textbf{Multi-Modality Segmentation.} Another line of research explored multi-modal segmentation, where models are trained using different imaging modalities such as CT, MRI, or PET, or by combining imaging with complementary clinical data \cite{wang2025difusionseg,zubair2025comprehensive, guo2019deep, albekairi2025multimodal, ren2021comparing, pichler2008positron, hossain2025enhancing,jafari2022lmisa,liao2025benchmarking}. These approaches aim to provide richer and more diverse\begin{algorithm}[H]

\caption{Joint Retrieval-Augmented Segmentation}
\label{alg:Joint_training}
\begin{algorithmic}[1]

 \STATE \textbf{Input:} dataset $\mathcal{D}$, Retrieval model $R_\theta$, Segmentation model $S_\phi$
 
\STATE Finetune $R_\theta$ and $S_\phi$ \com{// Independently}

\FOR{each epoch}

 \STATE  $\mathcal{D'}$ = $R_\theta$($\mathcal{D}$)  \com{// Knowledge base Embedding}
   
    \FOR{$(x, y)$ in $\mathcal{D}$}
    
        \STATE $q = R_\theta(x)$ \com{// Query image embedding}

        \STATE  $T_1, T_2 = \texttt{$Retrieve$}(q, \mathcal{D'})$ \com{// Top-k}

        \STATE $I_g, M_g = w_1 \cdot T_1 + w_2 \cdot T_2$ \com{// Guide Image/Mask}

        \STATE  $x' = \texttt{concat}(x, I_g, M_g)$ \com{// Query image + Guide }

        \STATE $\hat{y} = S_\phi(x')$ \com{// Prediction}
        
        \STATE $\mathcal{L} = \texttt{Loss}(\hat{y}, y)$ \com{// Loss}
        
        \STATE Update $R_\theta$ , $S_\phi$ \com{// Update Models}
       
    \ENDFOR
\ENDFOR
\end{algorithmic}
\end{algorithm} information to improve segmentation accuracy. However,
 they often require all modalities to be available for each patient, which is not always feasible, and differences in resolution, acquisition protocols, and noise characteristics can make integration challenging \cite{liao2025benchmarking, tan2025robust}. 

\vspace{5pt}
\textbf{Multi-Atlas Segmentation (MAS).} represents another established strategy, particularly in medical imaging. In MAS, a set of labeled atlas images is aligned to a target image through image registration, and the propagated labels are fused to produce the final segmentation \cite{iglesias2015multi, wang2012multi, heckemann2006automatic}. While MAS leverages prior labeled cases, it can be computationally expensive and sensitive to registration errors \cite{sabuncu2010generative}, and its performance depends heavily on the similarity between the atlas set and the target case \cite{tang2013bayesian, aljabar2009multi}.

\vspace{5pt}
\textbf{Few-Shot Segmentation.} To address some of these limitations, few-shot segmentation methods replace the rigid registration in MAS with learned feature representations from manually provided support cases. Shaban et al. \cite{shaban2017one} introduced a one-shot framework where a support image conditions the segmentation of a query image. Wang et al. \cite{wang2019panet} proposed PANet, which aligns and refines class prototypes between support and query features. Tian et al. \cite{tian2020prior} presented PFENet, which generates a prior mask from support examples to enrich query features. While effective, these methods relied on manually provided support images and kept support selection separate from segmentation training, limiting the potential for joint feature optimization between retrieval and segmentation.

\vspace{5pt}
\textbf{Retrieval Few-Shot Segmentation.}
Closest to our work is the retrieval-augmented few-shot framework \cite{zhao2025retrieval}, which extracts query embeddings, retrieves annotated examples from a memory bank, and conditions the segmentation model on them using the Segment Anything Model’s (SAM) \cite{ravi2024sam}, all without additional training. 
While effective for guidance, the retrieval module remains fixed as a similarity-based feature extractor, so it cannot adapt if retrieved samples fail to improve segmentation. 
Both retrieval and segmentation networks are kept frozen rather than independently or jointly optimized. 
Because SAM is a large foundation model, using it can pose substantial computational and deployment requirements, which may limit its immediate applicability in resource-constrained medical imaging environments.

\vspace{5pt}
These limitations motivate our proposed framework, a unified approach that jointly optimizes retrieval and segmentation within a single learning loop. Unlike prior methods that rely on a fixed, similarity-based retrieval stage, our retrieval network is trained through the segmentation loss itself, learning to select guidance images and masks that are functionally useful for delineating target structures rather than merely visually similar.
By integrating these dynamically retrieved examples into the segmentation pipeline, the model leverages contextual knowledge from similar past cases to enhance anatomical understanding and boundary precision. This joint optimization fosters mutual adaptation between retrieval and segmentation, reduces dependence on large labeled datasets, and delivers robust performance.

\vspace{-5pt}

\section{Methodology}
\vspace{-5pt}
\textbf{J}oint \textbf{R}etrieval \textbf{A}ugmented \textbf{S}egmentation (J-RAS) jointly trains a segmentation model and a retrieval model in a unified optimization loop. Retrieved image–mask pairs provide contextual guidance to enhance anatomical understanding, while the retrieval model is updated via segmentation loss to learn task-specific features. This mutual adaptation transforms retrieval from a passive reference into an active collaborator that boosts boundary precision and generalization. The overall framework is shown in \autoref{fig:pipeline}, and the training process in Algorithm\ref{alg:Joint_training}.

\subsection{Independent Optimization}
\vspace{-4pt}
Before integrating the models into the J-RAS framework, each component is first optimized independently.

\vspace{5pt}

\noindent
\textbf{Contrastive Training.} The retrieval encoder is fine-tuned using a contrastive learning objective, as illustrated in \autoref{fig:pipeline} (A.1). A multi-layer perceptron (MLP) projects the high-dimensional encoder output into a lower-dimensional embedding space suitable for similarity computation. Positive pairs are constructed from consecutive slices within the same patient volume end-diastolic (ED) or end-systolic (ES), while negative pairs are formed by sampling slices from different patients. Each pair is labeled as $y=1$ for positive and $y=0$ for negative. Training is performed with the Normalized Temperature-Scaled Cross-Entropy (NT-Xent) loss \cite{chen2020simple}, defined as:
\begin{equation}
\mathcal{L}_{\text{Contrastive}} = \frac{1}{2B} \sum_{i=1}^{2B} \text{CE}\Big([s_{i,p}, s_{i,n_1}, \dots, s_{i,n_{2B-2}}], 0 \Big),
\label{eq:ret_loss}
\end{equation}
where $B$ is the number of original samples in a mini-batch, yielding $2B$ augmented views. $s_{i,p}$ is the cosine similarity between the $i$-th sample and its positive pair, while $s_{i,n_j}$ are similarities to negatives in the batch. CE denotes cross-entropy loss over the similarity scores, with label $0$ for the positive pair. All similarities are scaled by a temperature parameter, while augmentations such as rotation, flipping, and contrast adjustment introduce additional positives, encouraging the embeddings to emphasize anatomical structures and remain robust to non-essential variations.

\vspace{5pt}
\noindent
\textbf{Supervised Training.} The segmentation model is fine-tuned after image normalization and standard augmentations, as illustrated in \autoref{fig:pipeline} (A.2). Training uses a combination of Dice loss and pixel-wise cross-entropy:

\begin{equation}
\mathcal{L}_{\text{Supervised}} = (1.0 - \text{Dice}_{\text{mean}}) + \text{CE}
\label{eq:seg_loss}
\end{equation}
where $\text{Dice}_{\text{mean}}$ is the mean Dice coefficient across all classes, and CE denotes cross-entropy loss. This stage ensures that the segmentation model learns accurate pixel-wise predictions before being incorporated into the joint retrieval-augmented training method.

\subsection{Joint Optimization}
After completing the independent optimization phase, the models are integrated into the J-RAS method. The optimization procedure proceeds in the following steps.

\vspace{5pt}
\noindent
\textbf{Retrieval Guides.}
At each training step, J-RAS passes the input image to the retrieval model, which computes cosine similarity against a set of dataset slices (Knowledge base) and selects the top-$k$ most similar images.
To ensure cross-patient guidance and avoid trivial matches, slices from the same patient as the query are excluded from retrieval. The top-$k$ retrieved images, along with their ground-truth segmentation masks, serve as contextual guides for the subsequent segmentation process.

\vspace{6pt}
\noindent
\textbf{Input Fusion.}
After retrieval, the segmentation model receives the query image and the top-$k$ guidance samples. To avoid overloading the network, these guides are merged into a single composite image and mask using temperature-scaled softmax weights:
\begin{equation}
w_i = \frac{\exp(sim_i / \tau)}{\sum_{j=1}^{k} \exp(sim_j / \tau)},
\end{equation}
where \(sim_i\) is the cosine similarity to the query and \(\tau\) is the temperature. The fused guidance pair \((\tilde{I},\tilde{M})\) is computed as:
\begin{equation}
\tilde{I} = \sum_{i=1}^{k} w_i I_i, \qquad
\tilde{M} = \sum_{i=1}^{k} w_i M_i.
\end{equation}
More similar samples contribute more strongly to the final guide. The query image (3 channels), fused guide image (3 channels), and mask (1 channel) are concatenated into a 7-channel tensor and projected via a lightweight adapter to the model’s 3-channel input space. This fusion integrates contextual cues and anatomical priors, improving delineation in ambiguous regions.

\vspace{6pt}
\noindent
\textbf{Mutual Adaptation.} To ensure mutual adaptation, the retrieval and segmentation models are jointly optimized at each training step using a shared segmentation loss (\autoref{eq:seg_loss}). This allows the retrieval model to learn representations increasingly aligned with the segmentation task, while the segmentation network benefits from semantically relevant guidance retrieved in real time.

The overall objective combines Dice and Cross-Entropy losses between the predicted and ground-truth masks, without any additional retrieval-specific term,\begin{table}[ht]
\centering
\resizebox{0.46\textwidth}{!}{%
\begin{tabular}{lccc}
\toprule
\textbf{Statistic} & \textbf{Split} & \textbf{ACDC} & \textbf{M\&Ms} \\
\midrule
\multirow{3}{*}{\#Patients} 
 & Training   & 100 & 150 \\
 & Validation & 0   & 34  \\
 & Testing    & 50  & 136 \\
\midrule
\multirow{3}{*}{\#Slices$_{\text{ED+ES}}$}
 & Training   & 1902 & 3286 \\
 & Validation & 0    & 806 \\
 & Testing    & 1076 & 3242  \\
\bottomrule
\end{tabular}}
\caption{Statistics of the ACDC and M\&Ms datasets. ED denotes the end-diastolic phase and ES the end-systolic phase.}
\vspace{-10pt}
\label{tab:datasetss}
\end{table} since the retrieval model has already learned to distinguish images based on generic similarity during the independent optimization phase. Although the loss remains unchanged, the concatenation of retrieved images and masks alters the input distribution, encouraging the segmentation model to fuse local query features with global priors from similar cases. This implicitly drives the retrieval branch to learn task-aware, semantically meaningful features that enhance segmentation accuracy.

The retrieval model is updated through the query branch to accelerate training. At the start of each epoch, all knowledge-base images are encoded once without gradients to obtain detached gallery embeddings that remain fixed throughout that epoch. For each query image, a trainable query embedding is computed and compared with the gallery embeddings to select the Top-\textit{K} guides. The segmentation loss then backpropagates through the similarity and weighting computations into the query embedding, updating the retrieval model. To keep the guidance relevant without increasing memory cost, the gallery embeddings are recomputed at the beginning of each epoch to stay aligned with the evolving retrieval model.

\section{Experimental Setup}
In this section, we describe the datasets, models, and evaluation metrics used to evaluate the effectiveness of the proposed method.

\subsection{Datasets}
We evaluate the Joint Retrieval-Augmented Segmentation (J-RAS) method on \textbf{two} publicly available datasets (\autoref{tab:datasetss}), which both focus on segmenting the same anatomical structures, enabling a controlled and meaningful assessment of cross-dataset generalizability.

\vspace{3pt}
\textbf{ACDC.} The Automated Cardiac Diagnosis Challenge dataset \cite{bernard2018deep} contains CMR scans from 150 patients acquired with multiple scanners and protocols, along with expert annotations for the right ventricle (RV), myocardium (MYO), and left ventricle (LV). For each patient, volumes include slices from end-diastolic (ED) and end-systolic (ES) phases, providing two distinct frames that capture different points in the cardiac cycle. As shown in \autoref{tab:datasetss}, the dataset is split into 100 patients for training and 50 for testing. Within the training set, we use an 80/20 split to create training and validation subsets. All images are resampled to a fixed resolution and normalized to ensure patient consistency. Performance is reported on the held-out test set, which comprises 50 patients, each with ED and ES frames, yielding a total of 100 test cases.

\vspace{7pt}
\textbf{M\&Ms.} The Multi-Centre, Multi-Vendor, and Multi-Disease Cardiac Image Segmentation Challenge dataset \cite{campello2021multi} includes 320 patients, as shown in \autoref{tab:datasetss}, scanned across multiple centers, vendors (Siemens, GE, Philips, Canon), and pathologies, introducing significant variability. The training set contains 150 annotated cases from three vendors, 34 cases for validation, and the 136-case test set includes studies from each known vendor plus studies from a fourth, previously unseen vendor to evaluate cross-domain generalization. We select M\&Ms because they target the same anatomical structures as ACDC, enabling assessment of J-RAS’s cross-dataset generalizability under realistic domain shifts.

\subsection{Models Evaluation}
\vspace{-3pt}
To evaluate our method, we employed several state-of-the-art segmentation architectures of varying complexity, including U-Net \cite{ronneberger2015u}, TransUNet \cite{chen2021transunet}, SegFormer \cite{xie2021segformer}, and the Segment Anything Model (SAM) \cite{kirillov2023segment}. These models span classical CNNs, hybrid CNN–Transformer designs, and lightweight or foundation-level transformer-based architectures, providing a broad basis for assessing retrieval-guided segmentation. For retrieval, we used the DINOv2 \cite{oquab2023dinov2} vision transformer pretrained on radiological data (RAD-DINO) \cite{perez2025exploring}, which captures fine-grained anatomical structures and modality-specific patterns, enabling semantically and anatomically relevant guidance without labeled supervision.

\subsection{Evaluation Metrics}
We evaluate segmentation performance using the Dice Similarity Coefficient (DSC) \cite{dice1945measures} and Hausdorff Distance (HD) \cite{huttenlocher2002comparing}. DSC measures overlap between predicted and ground-truth masks, while HD captures maximum boundary discrepancy, together providing a comprehensive assessment of accuracy and spatial error.

\section{Results and Discussion}
\vspace{-5pt}
In this section, we present a comprehensive evaluation of the proposed Joint Retrieval-Augmented Segmentation (J-RAS) method across multiple segmentation architectures. The analysis highlights the effectiveness of J-RAS in improving segmentation performance, while also providing both quantitative and qualitative insights into its behavior under different experimental settings.

\subsection{Quantitative Results}\label{Quantitative}
\vspace{5pt}
\subsubsection{Main Results}\label{sec:main_res} \autoref{tab:segmentation_results} summarizes the performance of the proposed J-RAS method at \textit{Top-$k$ = 2} (\autoref{Top-K-effect}) compared to using the segmentation models alone on the ACDC dataset. Results are reported using both Dice score and Hausdorff distance across the three anatomical classes. Overall, integrating J-RAS consistently improves performance across all models and metrics. For example, U-Net without J-RAS achieves an average Dice of 0.85 and Hausdorff distance of 2.27, which improves to 0.903 and 1.29, respectively, when\begin{table*}[t]
\centering
\begin{tabular}{lcccccccc}
\toprule
& \multicolumn{4}{c}{\textbf{Dice (↑)}} & \multicolumn{4}{c}{\textbf{HD (↓)}} \\
\cmidrule(lr){2-5} \cmidrule(lr){6-9}
\textbf{Method} 
& RV & MYO & LV & Mean $\pm$ STD 
& RV & MYO & LV & Mean $\pm$ STD  \\
\midrule

SegFormer & 0.8589 & 0.8484 & 0.8484 & 0.8708 $\pm$ 0.042& 1.8215 & 2.0030 & 1.6144 & 1.8130 $\pm$ 2.49 \\

\quad +\textit{\textbf{J-RAS}} & 0.9050 & 0.8900 & \textbf{0.9395} & 0.9115 $\pm$ 0.031 & \textbf{1.2706} & \textbf{1.0307} & \textbf{1.1455} & \textbf{1.1489 $\pm$ 0.30} \\
\midrule

UNet & 0.8688 & 0.7991 & 0.9084 & 0.8588 $\pm$ 0.040 & 3.1267 & 1.9004 & 1.8112 & 2.2790 $\pm$ 4.62 \\

\quad +\textit{\textbf{J-RAS}} & 0.8992 & 0.8797 & 0.9313 & 0.9034 $\pm$ 0.033 & 1.3887 & 1.3195 & 1.1814 & 1.2965 $\pm$ 0.70\\

\midrule

TransUNet & 0.8676 & 0.8418 & 0.9190 & 0.8761 $\pm$ 0.035 & 2.3897 & 1.2355 & 1.2894 & 1.6382 $\pm$ 2.14\\

\quad +\textit{\textbf{J-RAS}} & \textbf{0.9105} & \textbf{0.8918} & 0.9330 & \textbf{0.9118 $\pm$ 0.030} & 1.9041 & 1.2598 & 1.7063 &1.6234 $\pm$ 3.30\\

\midrule

SAM & 0.6009 & 0.6416 & 0.7450 & 0.6625 $\pm$ 0.127 & 28.089 & 20.611 & 28.247 & 25.649 $\pm$ 20.6\\

\quad +\textit{\textbf{J-RAS}} & 0.8352 & 0.8450 & 0.9219 & 0.8674 $\pm$ 0.042 & 4.3056 & 2.8576 & 2.5400 &3.2344 $\pm$ 4.77\\

\midrule
\makecell{Retrieval-Augmented\\Few-shot \cite{zhao2025retrieval}} & 0.6729 & 0.7757 & 0.8472 & 0.7652 & -- & -- & -- & -- \\

\bottomrule
\end{tabular}
\vspace{-5pt}
\caption{Performance comparison of the segmentation models on the ACDC test set, evaluated both using the segmentation model alone and with the proposed method (\textit{J-RAS}) at \textbf{\textit{Top-$k$ = 2}}. Results are reported across three classes as well as their average. Dice ($\uparrow$) denotes the Dice score, where higher values indicate better performance, while HD ($\downarrow$) denotes the Hausdorff distance, where lower values indicate better performance. For completeness, we also include results from a retrieval-augmented few-shot method \cite{zhao2025retrieval}.}
\vspace{-5pt}
\label{tab:segmentation_results}
\end{table*} combined with J-RAS. Similarly, SAM shows marked gains, with Dice increasing from 0.66 to 0.86 and Hausdorff distance dropping from 25.649 to 3.234. Within the J-RAS, both TransUNet and SegFormer achieve similar mean Dice scores 0.9118 and 0.9115 respectively. However, SegFormer stands out as the best model by also attaining the lowest Hausdorff distance of 1.1489. 

This superior performance can be attributed to SegFormer’s strong global attention and efficient multi-scale feature representation without relying on heavy positional encodings. TransUNet attains comparable results but with higher computational complexity, whereas U-Net performs slightly worse due to its limited capacity for contextual modeling. SAM exhibits the weakest performance, as its complex yet general-purpose design lacks adaptation to the domain-specific intensity patterns and structural nuances of medical images.

\vspace{4pt}
\textbf{Comparison with Related Work.} To contextualize J-RAS performance, we consider both retrieval-based and segmentation-only baselines. First, we evaluate the recently proposed retrieval-augmented few-shot segmentation method by Zhao et al. \cite{zhao2025retrieval}, which is designed for few-shot segmentation without additional training and relies on DINOv2 and SAM models. As shown in \autoref{tab:segmentation_results}, on the ACDC dataset, it achieves a mean Dice score of 0.7652, lower than SAM within J-RAS (0.8674) and substantially lower than SegFormer/TransUNet within J-RAS 0.9115/0.9118. This gap highlights the benefit of jointly training retrieval and segmentation models rather than using a fixed retrieval module solely as a feature extractor without end-to-end training.

For segmentation-only baselines, we evaluated different backbones (U-Net, TransUNet, SegFormer, and SAM) to provide a comprehensive comparison. TransUNet\begin{figure}[ht]
\centering
\includegraphics[width=0.49\textwidth]{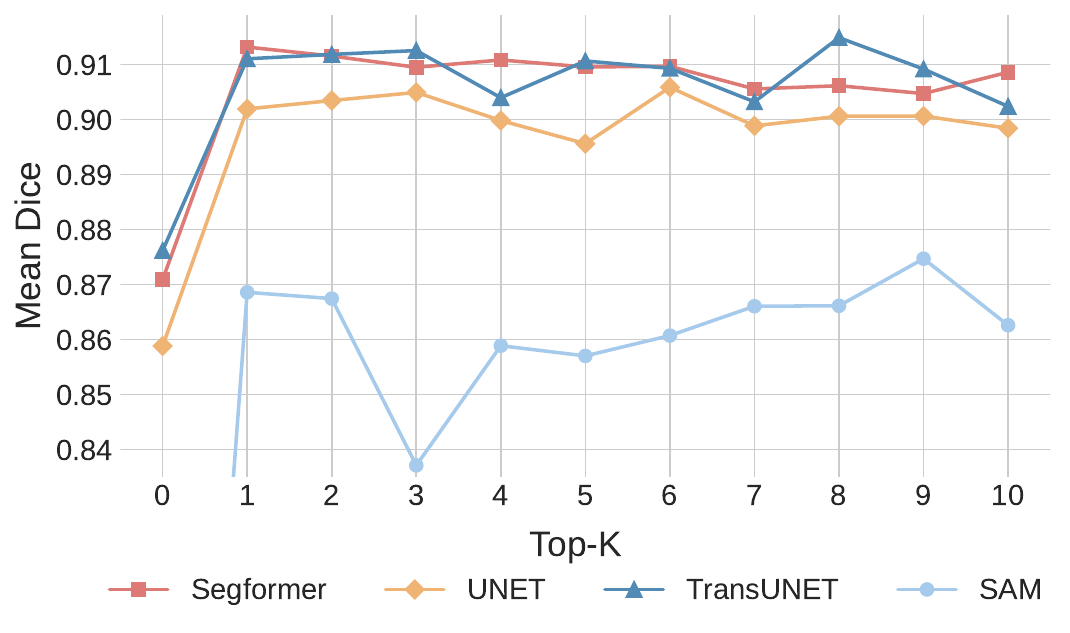}
\vspace{-20pt}
\caption{Mean Dice scores on the ACDC test set across different numbers of retrieved guide images and masks (\textit{Top-K = 1–10}). The case of \textbf{Top-K = 0} represents the baseline models without J-RAS, where SAM attains a score of 0.66.}
\label{fig:TopK_AVG_Dice}
\vspace{-10pt}
\end{figure} has also been evaluated in prior work on the ACDC dataset \cite{chen2021transunet}, despite not being retrieval-based. Including these baselines allows for a more precise comparison of J-RAS against both retrieval-based methods \cite{zhao2025retrieval} and segmentation-only architectures.

\subsubsection{Effect of Top-K Retrieval on Segmentation Performance}\label{Top-K-effect} We investigate the effect of varying the number of retrieved guide images and masks (\textit{Top-K}) on segmentation performance. \autoref{fig:TopK_AVG_Dice} reports the mean Dice scores across the three target structures for U-Net, TransUNet, SAM, and SegFormer. All models show consistent improvements when incorporating retrieval guidance, with performance increasing as $K$ grows from 1 to around 5–6. However, the gains beyond $K=2$ are modest, and at higher $K$ values performance tends to saturate or slightly decline, likely due to noise from less relevant images.

\begin{figure}
\centering
\includegraphics[width=0.49\textwidth]{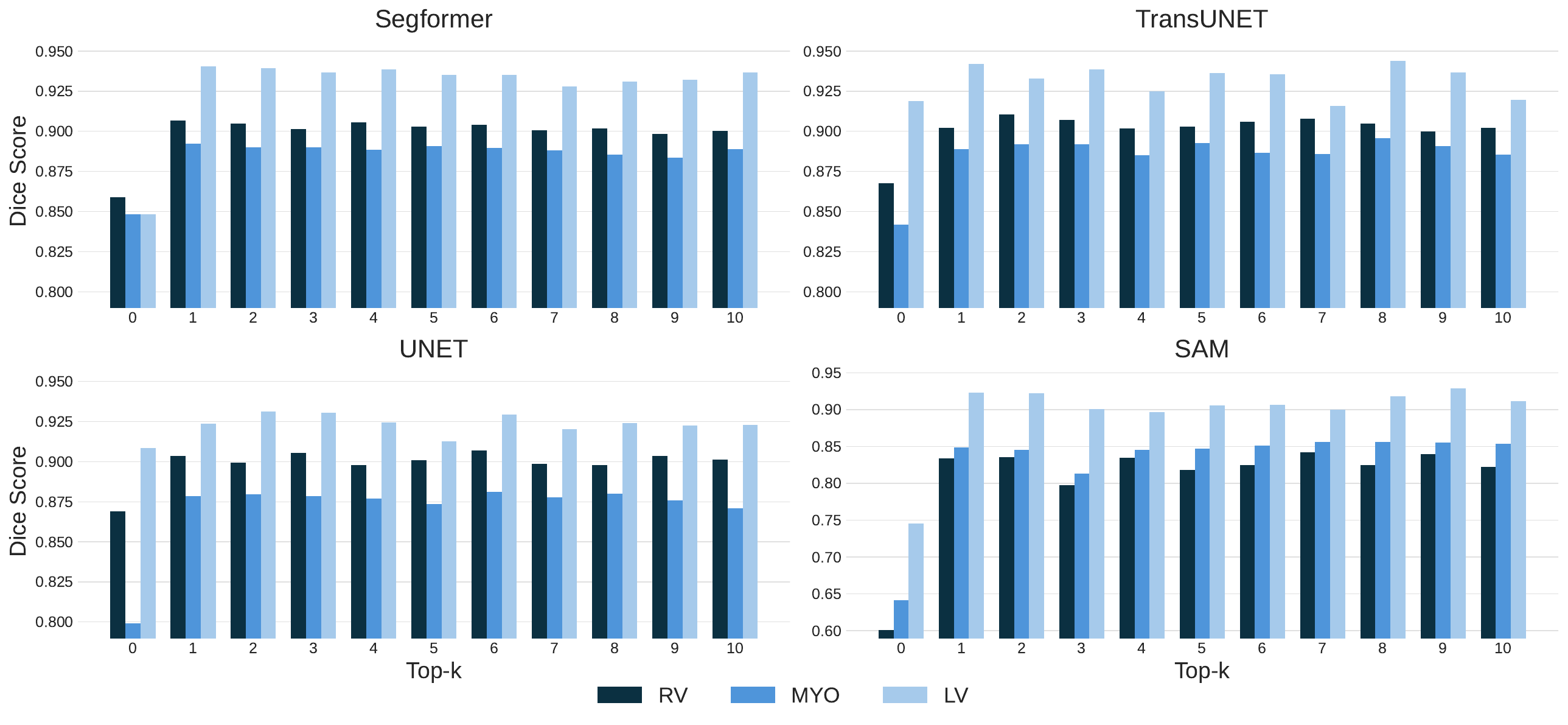}

\vspace{-10pt}
\caption{Dice scores for Classes (RV, MYO, and LV) using \textit{Top-K} retrieved images and masks ($K = 1$–$10$) on the ACDC test set with SegFormer, TransUNet, U-Net, and SAM. Bars represent class-wise performance. \textbf{K = 0} corresponds to the \textbf{baseline method}.
\vspace{-12pt}
}

\label{fig:per_class_dice}
\end{figure}

To better understand the class-wise behavior, we analyze the Dice scores for each anatomical structure separately (\autoref{fig:per_class_dice}). The results show that changes in $K$ do not uniformly affect all classes. For example, in U-Net, Class 3 achieves a higher Dice score at $K=2$ compared to $K=1$, while Class 1 performs better at $K=1$. This indicates that the optimal number of retrieved images can vary across classes, reflecting differences in structural complexity and retrieval relevance. Based on these findings, we fix $K=2$ for all experiments, as it provides a balanced trade-off without introducing noise.
\begin{figure}[ht]
\centering
\includegraphics[width=0.47\textwidth]{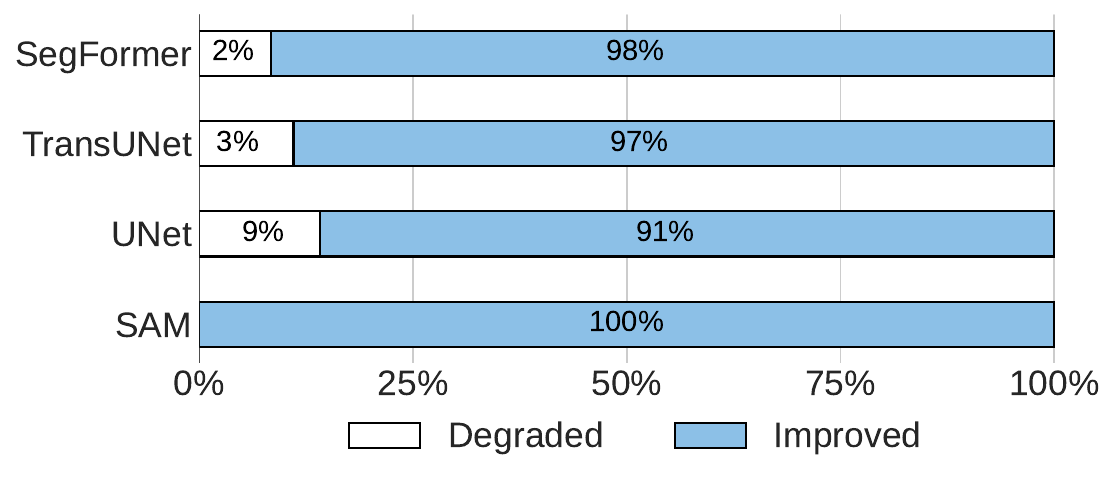}
\vspace{-14pt}
\caption{Case-level analysis of improvements and degradations on the ACDC test set using the J-RAS method with SegFormer, TransUNet, SAM, and U-Net. The number of patients whose Dice scores improved or degraded compared to the baseline segmentation models is shown on the respective segment.}
\vspace{-15pt}
\label{fig:degraded_enhanced_patients}
\end{figure}
\subsubsection{Case-Level Improvements and Degradations}\label{sec:case_analysis}
We further analyze per-case improvements and degradations to assess how the J-RAS method impacts individual patient results compared to the baseline models alone. \autoref{fig:degraded_enhanced_patients} reports the results on the ACDC test set comprising 100 patients. When integrated within J-RAS, SegFormer improved performance in 98\begin{figure}[ht]
\centering
\includegraphics[width=0.49\textwidth]{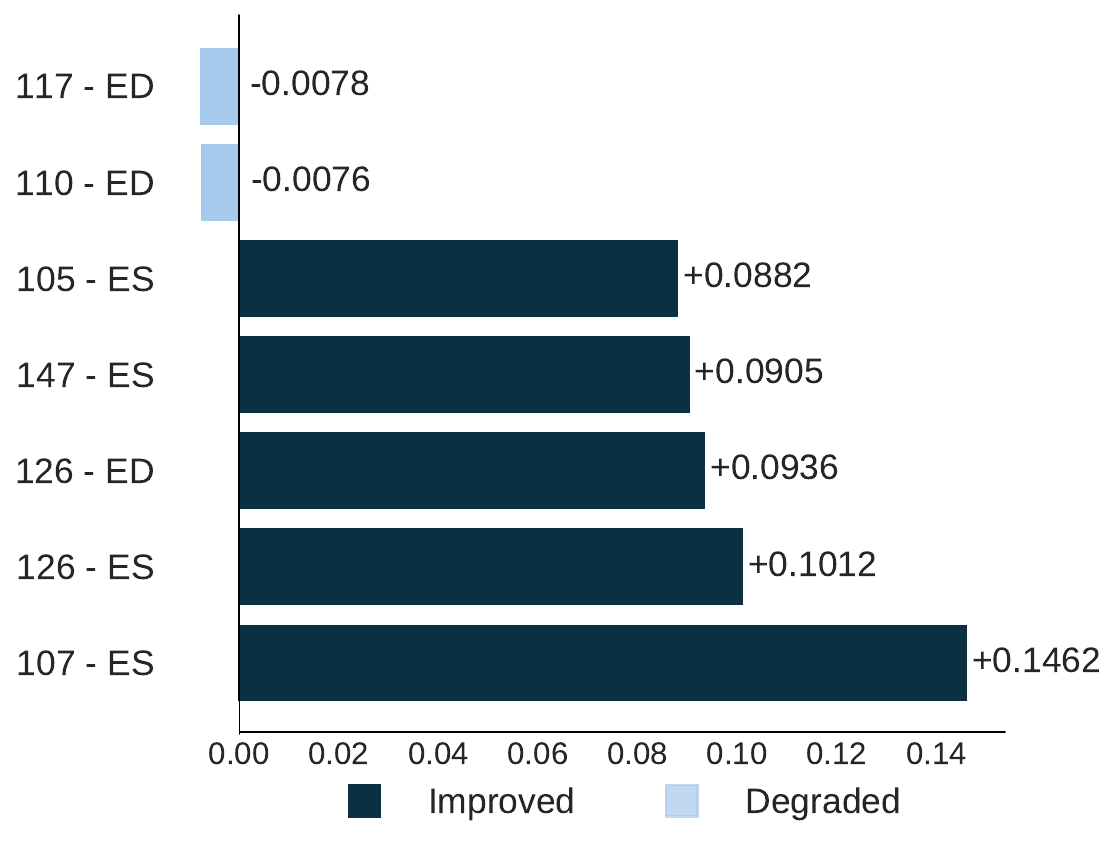}
\vspace{-25pt}
\caption{Top five most improved patients (out of 98) and the two degraded cases on the ACDC test set when using the J-RAS method with SegFormer. 
Bars indicate the magnitude of change in Dice score relative to the baseline model.}
\vspace{-6pt}
\label{fig:Top5improved_2degraded}
\end{figure} cases, with only 2 showing a slight degradation in Dice score. TransUNet improved 97 cases while 3 degraded, and U-Net improved 91 cases while 9 degraded. Notably, SAM achieved improvements in all 100 patients without any degradations.

To further quantify the magnitude of changes, we measured the degree of improvement and degradation for each patient, as illustrated in \autoref{fig:Top5improved_2degraded}. 
The figure highlights the top five most improved patients using J-RAS with SegFormer, as well as the two degraded cases. 
For instance, patient 107 (end-systolic frame) improved by \textit{+0.1462} Dice, and patient 126 (end-systolic frame) improved by \textit{+0.1012}. 
In contrast, the only degraded cases were patient 117 and patient 110 (both end-diastolic frames), with slight Dice reductions of \textit{-0.0078} and \textit{-0.0076}, respectively. Upon closer inspection, we found that these degradations often occurred when the retrieved guide images were not very close matches, largely due to poor slice quality in the query images. This reduced the effectiveness of the retrieval process and, in turn, \begin{figure*}
\centering
\includegraphics[width=1\textwidth]{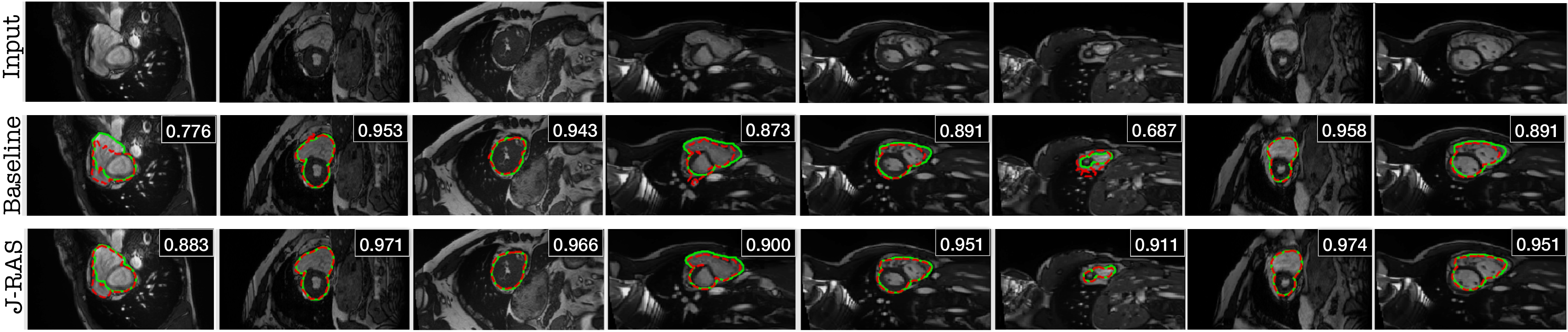}

  \vspace{-5pt}
  \caption{Qualitative segmentation results on the ACDC test set using \textbf{SegFormer} as the baseline and the proposed \textbf{J-RAS method} with SegFormer at \textbf{Top-$k=2$}. Each column shows an input image, with ground-truth contours overlaid (in green) and predictions (in red). Rows compare the baseline SegFormer (middle) against J-RAS (bottom). The \textbf{Dice scores} above each predicted mask highlight the improved segmentation performance achieved by J-RAS over the baseline.}
   \vspace{-10pt}
  \label{fig:contours}
\end{figure*} impacted segmentation accuracy. Nonetheless, these results confirm that J-RAS yields consistent per-case benefits overall with slight degradations.

\subsection{Qualitative Results}
\begin{figure}[ht]
\centering
\includegraphics[width=0.49\textwidth]{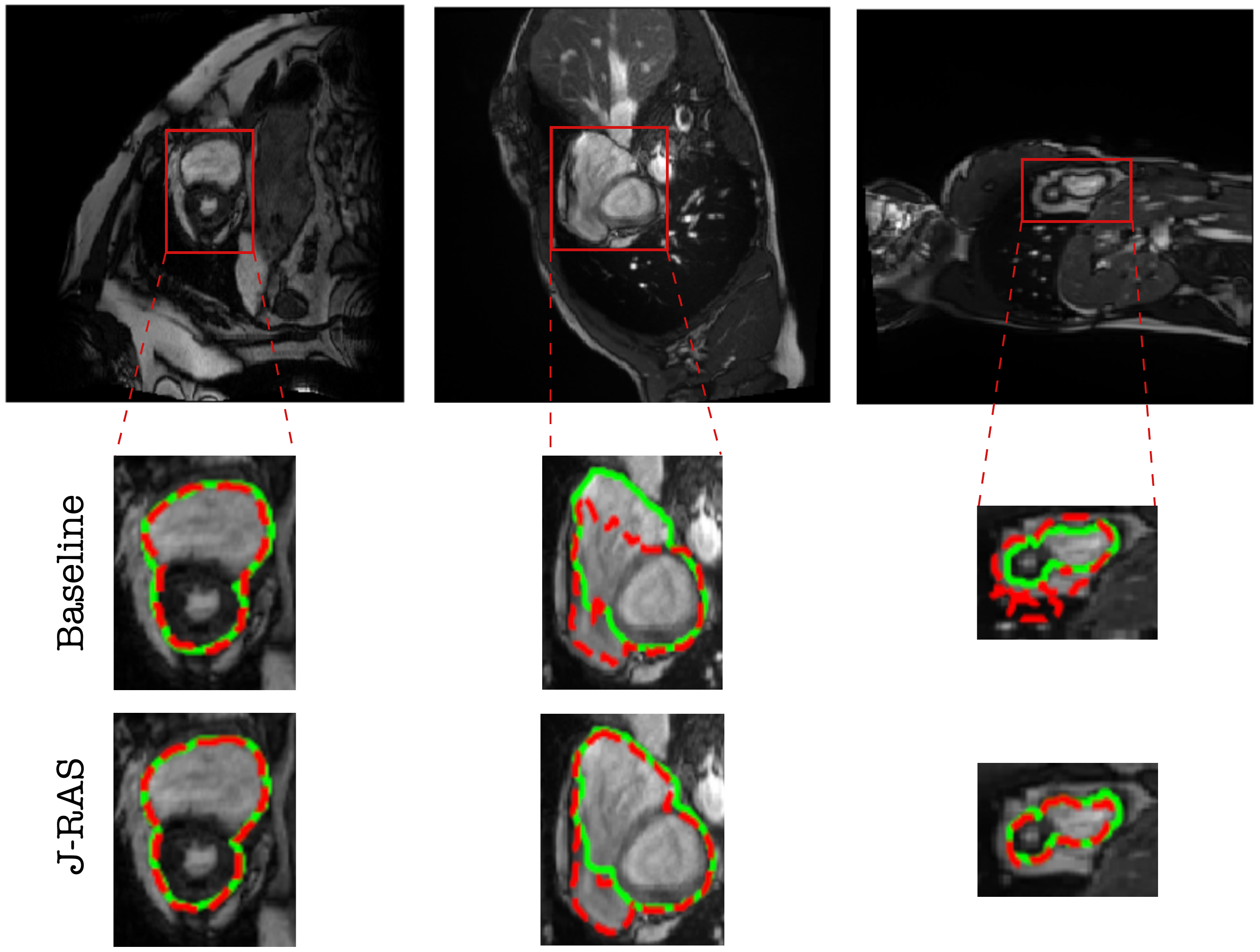}
\vspace{-18pt}
\caption{Comparison of segmentation results on samples from the ACDC test set. The top row shows the input images with the region of interest highlighted. For each case, the baseline segmentation (middle row) and the proposed J-RAS method (bottom row) are overlaid in comparison with ground-truth annotations (green). Red dashed lines denote model predictions.}
\label{fig:zoomed}
\vspace{-13.0pt}
\end{figure}

\begin{figure}[ht]
\centering
\includegraphics[width=0.48\textwidth]{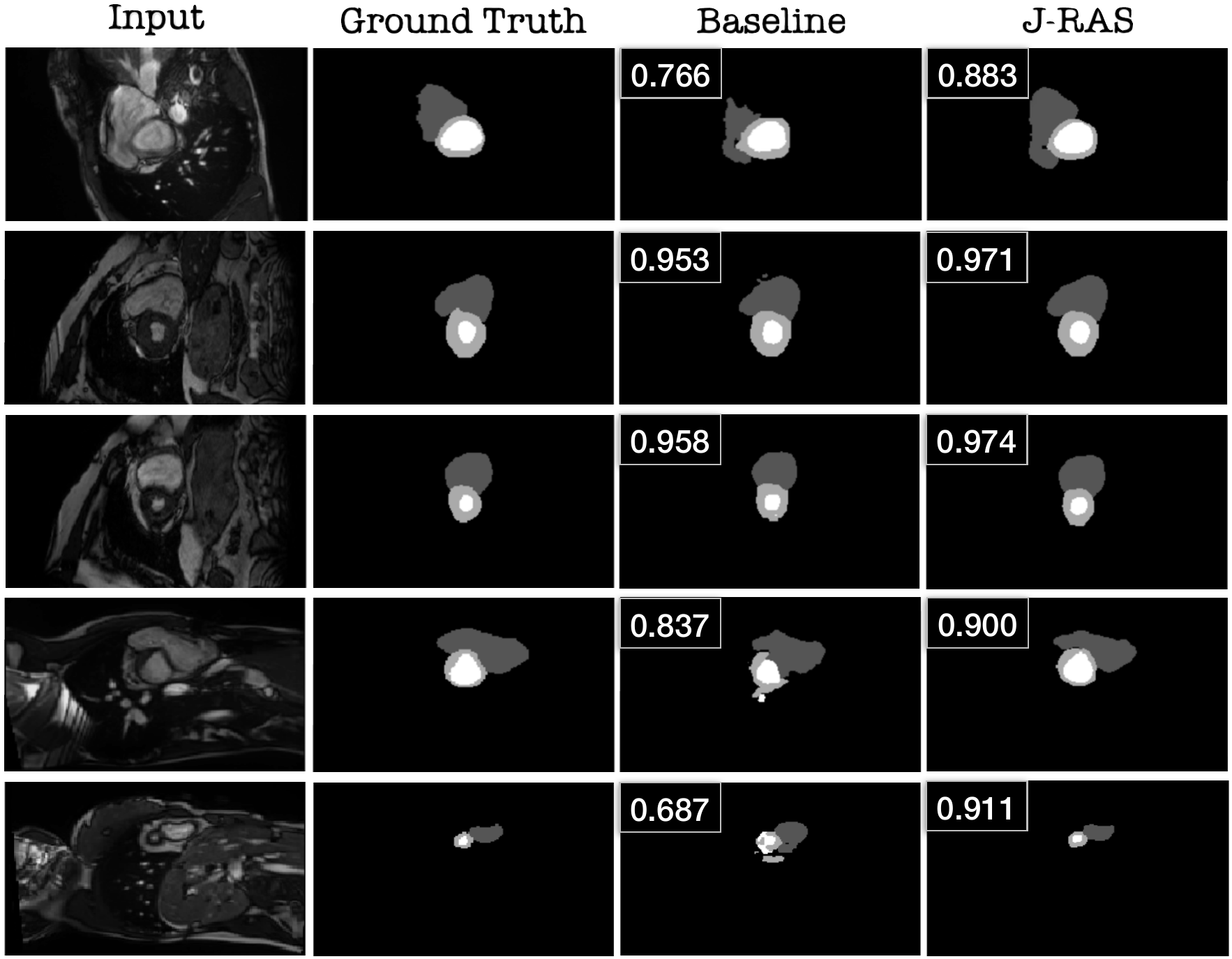}
\vspace{-5pt}
\caption{Comparison of segmentation results using \textbf{SegFormer} as the baseline and the proposed \textbf{J-RAS method} with SegFormer at \textbf{Top-$k=2$} on the ACDC test set. Each row shows the input image, ground truth mask, baseline SegFormer, and J-RAS prediction. The \textbf{Dice score} for each predicted mask is reported above the corresponding output, illustrating the performance improvement of J-RAS over the baseline.}
\vspace{-10pt}

\label{fig:pred_gt}
\end{figure}

\begin{figure}[ht]
\centering
\includegraphics[width=0.49\textwidth]{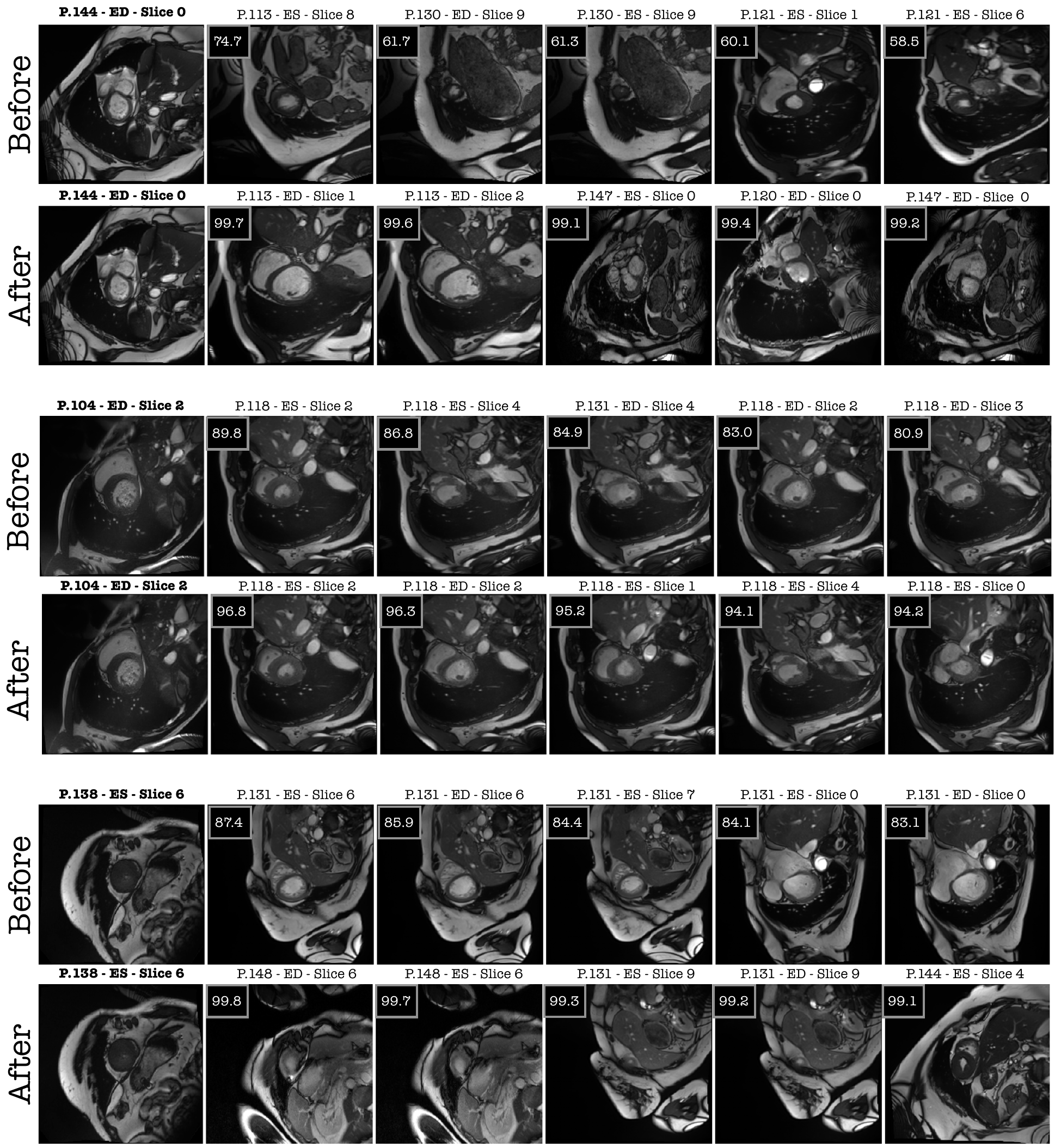}
\vspace{-18pt}
\caption{Comparison of retrieved image results using the retrieval model before and after the proposed \textbf{J-RAS method} on patients from the ACDC test set. The first image in each row, with a bold title, represents the query image, while the remaining images are the retrieved results. \textit{P.144} denotes patient 144; \textit{ES} indicates the end-systolic frame, and \textit{ED} indicates the end-diastolic frame. The cosine similarity between each retrieved image and the query image is shown in a black box at the top-left corner of each image.}
\vspace{-10.0pt}
\label{fig:ret_res}
\end{figure}

In this section, we qualitatively evaluate the effectiveness of the proposed J-RAS method by comparing its segmentation outputs against the baseline SegFormer model. Representative examples from the ACDC test set are presented in Figures (\ref{fig:contours}, \ref{fig:zoomed}, \ref{fig:pred_gt}, \ref{fig:ret_res}), with results shown at Top-$k=2$.

\vspace{5pt}
\textbf{Boundary Precision.} \autoref{fig:contours} highlights boundary precision by presenting representative input slices with ground-truth contours (green) and predicted contours (red). Compared to the baseline, J-RAS achieves tighter alignment with the ground truth, reducing misalignments and producing smoother, more reliable contours. The improvements are particularly evident in challenging cases with anatomical variability or weak contrast, where baseline predictions often fail to capture the true structure. To further illustrate these gains, \autoref{fig:zoomed} presents zoomed-in regions of interest, comparing SegFormer alone with SegFormer integrated within J-RAS. This view reveals finer details of the boundaries, showing that J-RAS consistently reduces boundary errors and better adheres to the true anatomy across different views. Even in cases where the baseline performs reasonably well, J-RAS enhances boundary fidelity and strengthens alignment with the ground truth, contributing to improved Dice scores.

\vspace{5pt}
\textbf{Mask-Level Improvements.} \autoref{fig:pred_gt} highlights improvements at the mask level, where each row shows the input image, the ground-truth mask, the baseline SegFormer prediction, and the J-RAS prediction. The Dice scores above each predicted mask demonstrate that J-RAS consistently yields more accurate segmentations. In challenging cases with small or low-contrast cardiac structures, the baseline often produces incomplete or fragmented predictions. At the same time, J-RAS generates more complete and coherent masks (e.g., improving Dice from 0.687 to 0.911). Even in cases where the baseline already performs well, J-RAS further refines the predictions, leading to consistent accuracy gains.

\vspace{5pt}
\textbf{Retrieval Results.} \autoref{fig:ret_res} illustrates the retrieval performance before and after applying the proposed J-RAS method. The results demonstrate that joint training substantially improves the ability of the retrieval model to identify slices that are anatomically and spatially closer to the query image. For example, in the top two rows, the query image is slice 0 from patient 144. Before joint training, the model retrieved slice 8 from patient 113 as the most similar, which is spatially distant 
from the query slice. After applying J-RAS, the model correctly identified slice 1 from the same patient as the closest match, which is much more consistent with the query slice. The cosine similarity scores shown above each retrieved image also increase significantly, further confirming the improved alignment. Additionally, a similar trend can be seen in the middle rows, where the query image is slice 2 from patient 104. Before J-RAS, the model mistakenly retrieved slice 4 from other patients multiple times, showing poor alignment. After joint training, the retrievals shifted to slices adjacent to the query (slices 0-3), with consistently higher cosine similarity values, reflecting a more anatomically accurate match.

Finally, in the bottom rows, the query image is slice 6 from patient 158. Before J-RAS, retrieved images were often rotated or flipped relative to the query. In contrast, after joint training, the model retrieved images that not only correspond to the same anatomical slice but also preserve the correct orientation, highlighting the robustness of the improved retrieval. 

These qualitative results demonstrate that retrieval-guided segmentation improves both overall mask accuracy and boundary alignment, reinforcing the robustness and generalizability of J-RAS across diverse anatomical and imaging conditions. These 
examples also confirm the potential of J-RAS to enhance retrieval consistency, spatial coherence, and anatomical alignment, ultimately providing more meaningful guidance for segmentation. These visual improvements are consistent with the quantitative results reported in \autoref{Quantitative}, where J-RAS achieves higher Dice scores and lower Hausdorff distances across multiple classes compared to the baseline.

\vspace{-10pt}
\section{Ablation Studies}
\vspace{-10pt}
In this section, we evaluate the robustness and effectiveness of the proposed J-RAS method through a series of ablation studies and analyses.
\subsection{Fusion Strategies}\label{sec:fusion}
We investigate strategies to fuse guidance information (retrieved images and masks) with the query image to improve segmentation. Since J-RAS provides\begin{figure*}
\centering
\includegraphics[width=1\textwidth]{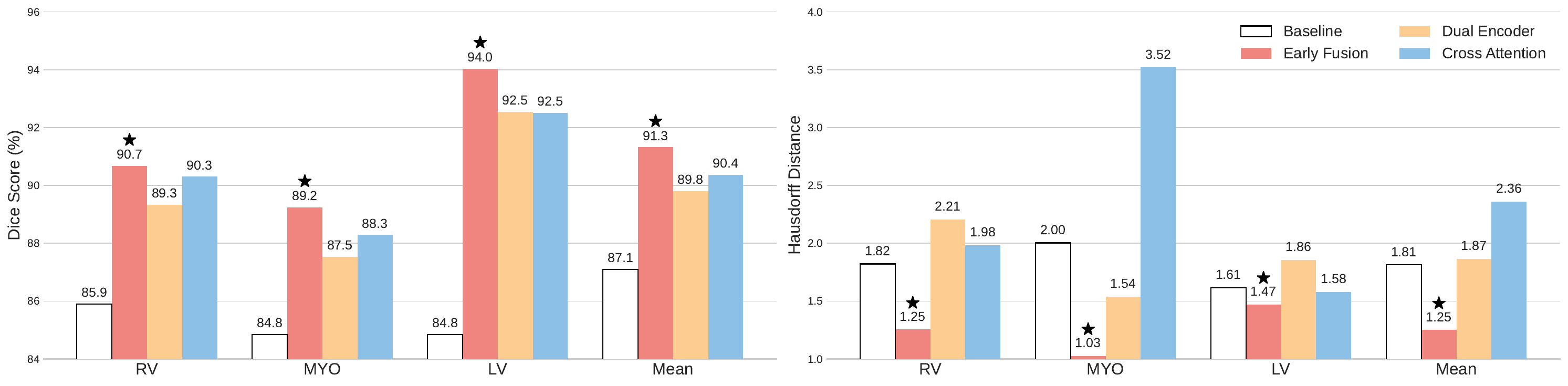}
\vspace{-18pt}
\caption{Comparison of fusion strategies in the J-RAS method on the ACDC test set. Results are reported using Dice score (↑, higher is better) and Hausdorff Distance (↓, lower is better). The methods include early fusion, cross-attention fusion, and a dual-encoder fusion approach. A $\bigstar$ above a bar indicates the best performance for the corresponding metric. Overall, early fusion achieved the highest Dice score and the lowest Hausdorff Distance, outperforming the other fusion methods.}
\vspace{-10pt}
\label{fig:fusion_methods}
\end{figure*} complementary structural and contextual cues, the fusion mechanism is crucial for effective use by the segmentation backbone. Using SegFormer at \textit{topk=1}, we implement and compare three approaches that differ in how deeply the guidance is integrated into the query representation.

\vspace{5pt}
\textbf{Early Fusion.}  
In early fusion, the query image, guide image, and guide mask are concatenated along channels, then projected to three channels via a lightweight $1\times1$ convolution. Early fusion injects guide information at the pixel level, enabling the network to align query and guide features from the first layer.

\vspace{5pt}
\textbf{Cross-Attention Fusion.} In this approach, a cross-attention module fuses query and guide features, with query features as queries and guide features as keys and values. This allows the model to focus on guide regions relevant to ambiguous or low-contrast query areas, offering interpretable interactions at the cost of higher computation.

\vspace{5pt}
\textbf{Dual-Encoder Fusion.} In this design, query and guide images are processed by separate encoders, with the guide mask modulating guide features to emphasize relevant anatomy. Encoded outputs are merged at an intermediate level for joint decoding, preserving independent representations but increasing model size and training cost.

\vspace{5pt}
The results in \autoref{fig:fusion_methods}  highlight the trade-offs of each fusion design. Early fusion achieves the highest overall performance (Dice 91.3, HD 1.25), indicating that pixel-level integration consistently benefits segmentation. Cross-attention surpasses dual-encoder in Dice (90.4 vs. 89.8), capturing fine-grained structural correspondences,\begin{table}
\centering
\resizebox{0.48\textwidth}{!}{ 
\begin{tabular}{lcccc}
\toprule
 & \multicolumn{4}{c}{\textbf{Dice (↑)}} \\
\cmidrule(lr){2-5}

\textbf{Top-$k$ Strategy} 
& RV & MYO & LV & Mean $\pm$ STD\\
\midrule

Fixed Top-$k$   & \textbf{0.9067} & \textbf{0.8924} & \textbf{0.9404} & \textbf{0.9132 $\pm$ 0.028}  \\

Dynamic Top-$k$ & 0.9016 & 0.8869 & 0.9275 & 0.9053 $\pm$ 0.035 \\

\midrule
 & \multicolumn{4}{c}{\textbf{HD (↓)}} \\
\cmidrule(lr){2-5}
\textbf{Top-$k$ Strategy} & RV & MYO & LV & Mean $\pm$ STD \\
\midrule

Fixed Top-$k$   & \textbf{1.2548} & \textbf{1.0266} & \textbf{1.4702} & \textbf{1.2505 $\pm$ 1.35} \\

Dynamic Top-$k$ & 1.8852 & 1.1329 & 1.7691 & 1.5957 $\pm$ 2.67\\
\bottomrule
\end{tabular}}
\vspace{-4pt}
\caption{Comparison of Fixed and Dynamic Top-$k$ retrieval strategies within the J-RAS method on the ACDC test set. Results are reported using Dice score (↑, higher is better) and Hausdorff Distance (HD, ↓, lower is better).}
\vspace{-14pt}
\label{tab:dynamictopk}
\end{table} while dual-encoder yields lower HD (1.87 vs. 2.36), preserving boundary accuracy. Overall, these findings suggest that although advanced mechanisms offer targeted advantages, early fusion remains highly effective with a strong backbone like SegFormer, underscoring the need to balance architectural complexity with performance gains when using retrieval-augmented guidance.

\subsection{Dynamic Top-K Retrieval} \label{sec:DynamicK}
In our experiments with the J-RAS method, we observe that using a fixed $k$ did not consistently improve results across all classes or cases. In some instances, performance increased, while in others it decreased. To address this, we explored a dynamic Top-$k$ vs fixed Top-$k$ strategy within the method as shown in \autoref{tab:dynamictopk}, using SegFormer as the segmentation backbone. Instead of predefining $k$ as a fixed number, this method adaptively determines the number of retrieved samples for each query based on the\begin{table*}[t]
\centering

\begin{tabular}{lcccccccc}

\toprule
\multirow{2}{*}{\textbf{Method}} &
\multicolumn{2}{c}{\textbf{RV}} &
\multicolumn{2}{c}{\textbf{MYO}} &
\multicolumn{2}{c}{\textbf{LV}} &
\multicolumn{2}{c}{\textbf{Mean $\pm$ STD}} \\
\cmidrule(lr){2-3} \cmidrule(lr){4-5} \cmidrule(lr){6-7} \cmidrule(lr){8-9}
 & {\makecell{\textbf{Dice} $\uparrow$}} & {\makecell{\textbf{HD} $\downarrow$}}
 & {\makecell{\textbf{Dice} $\uparrow$}} & {\makecell{\textbf{HD} $\downarrow$}}
 & {\makecell{\textbf{Dice} $\uparrow$}} & {\makecell{\textbf{HD} $\downarrow$}}
 & {\makecell{\textbf{Dice} $\uparrow$}} & {\makecell{\textbf{HD} $\downarrow$}} \\
\midrule

\textbf{SegFormer}    
&   0.8589 & 
\multicolumn{1}{c}{1.8215}
& 0.8484 & \multicolumn{1}{c}{2.0030} 
&   0.8484 & \multicolumn{1}{c}{1.6144} 
&   0.8708 $\pm$ 0.042 & \multicolumn{1}{c}{1.8130 $\pm$ 2.49} \\
\addlinespace[2pt]

\textit{\textbf{+ J-RAS}} 
& 0.9067 & 1.2548 
& \textbf{0.8924} & \textbf{1.0266}
& \textbf{0.9404} & \textbf{1.4702}
& \textbf{0.9132 $\pm$ 0.028} & \textbf{1.2505 $\pm$ 1.35}\\

\textit{\textbf{+ J-RAS}$_{\text{Gaussian}}$}
& 0.8984 & 1.3467
& 0.8757 & 1.7341
& 0.9140 & 1.8724
& 0.8960 $\pm$ 0.039 & 1.6511 $\pm$ 1.89\\

\textit{\textbf{+ J-RAS}$_{\text{SaltAndPepper}}$}
& \textbf{0.9071} & \textbf{1.2507}
& 0.8830 & 1.5406
& 0.9241 & 1.5045
& 0.9047 $\pm$ 0.033 & 1.4319 $\pm$ 2.63 \\

\textit{\textbf{+ J-RAS}$_{\text{Dropout}}$} 
& 0.8975 & 1.3501
& 0.8814  & 1.3708
& 0.9347 & 1.1725
& 0.9045 $\pm$ 0.035 & 1.2978 $\pm$1.13 \\

\bottomrule
\end{tabular}

\caption{Evaluation of the J-RAS method with SegFormer on the ACDC test set at \textbf{Top-$k=1$}. The experiments assess the impact of introducing different noise types into the guide images and masks. Results are reported using the Dice score ($\uparrow$ higher is better) and Hausdorff distance HD ( $\downarrow$ lower is better).}
\vspace{-8pt}
\label{tab:segmentation_results_noise}
\end{table*} \begin{figure}[ht]
\centering
\includegraphics[width=0.48\textwidth]{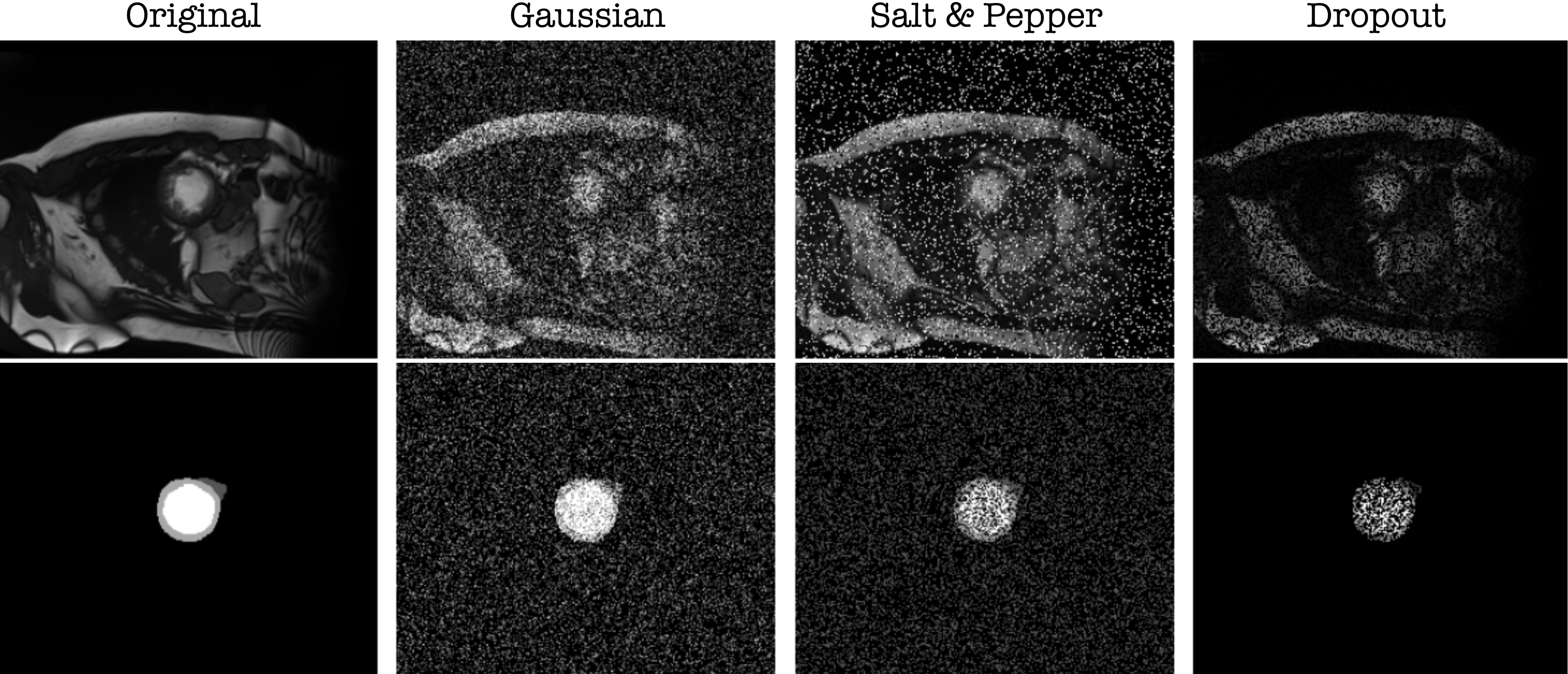}
\vspace{-8pt}
\caption{Illustration of different noise types applied to the guide images/masks. From left to right: the original image, Gaussian noise, salt-and-pepper noise, and dropout noise.}
\vspace{-11pt}
\label{fig:Noiseee}
\end{figure} similarity distribution of candidate knowledge base images. Specifically, for each query, we compute cosine similarity scores against all knowledge base embeddings and then count the number of retrieved samples whose similarity exceeds a predefined threshold $\theta$. The effective Top-$k$ for that query is dynamically set as:

\begin{equation}
k = \max \left( k_{\min}, \; \min(\text{count}_{\text{above-}\theta}, \; k_{\max}) \right)
\end{equation}

Here, $k_{\min}$ and $k_{\max}$ bound the minimum and maximum number of retrieved samples, which we set to 1 and 10, respectively. This ensures that at least one guide is always available while avoiding excessive retrieval when only a few samples are sufficiently similar.

\noindent
This adaptive approach allows the retrieval component to flexibly adjust the guidance depending on the available information. However, our experiments revealed that the dynamic Top-$k$ strategy did not yield significant improvements. using a fixed Top-$k$ with $k=1$ achieved better performance, with a mean Dice score of 0.9132 and a mean HD of 1.2505, compared to the dynamic strategy, which achieved a mean Dice score of 0.9053 and a mean Hausdorff Distance of 1.5957.
\begin{figure}[ht]
\centering
\includegraphics[width=0.48\textwidth]{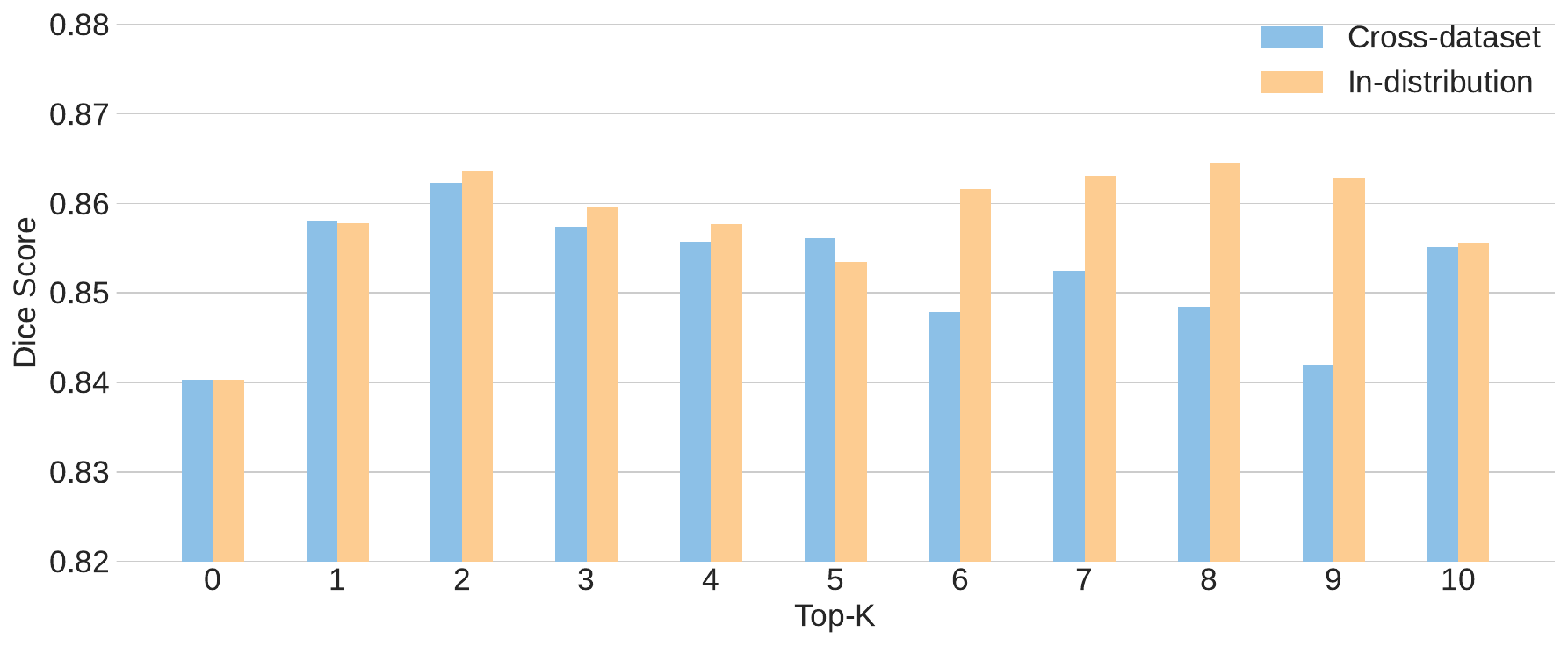}
\vspace{-10pt}
\caption{Mean Dice scores on the M\&Ms dataset using the J-RAS method under two settings: (i) \textbf{cross-dataset retrieval}, where segmentation is performed on M\&Ms and guides are retrieved from ACDC, and (ii) \textbf{in-distribution retrieval}, where both segmentation and retrieval are performed on M\&Ms. Results are reported across Top-$K$ values ($K = 1$--$10$), with $K = 0$ corresponding to the baseline model without retrieval.}
\vspace{-12pt}
\label{cross_in}
\end{figure} 

\subsection{Robustness to Noisy Guides} \label{sec:Noisee}
We evaluated the robustness of J-RAS by introducing noise into the retrieved guide images and masks (\autoref{fig:Noiseee}) and compared performance against two baselines, SegFormer without J-RAS and SegFormer within J-RAS using clean guides. Three types of noise were applied, Gaussian, salt-and-pepper (SP), and dropout. As shown in \autoref{tab:segmentation_results_noise}, clean guides yielded the best results (Dice 0.913, HD 1.25). Adding noise led to slight performance drops, with Dice/HD scores of 0.8960/1.651 for Gaussian, 0.9047/1.431 for SP, and 0.9045/1.297 for dropout. Despite these reductions, J-RAS with noisy guides still outperformed the baseline SegFormer model, demonstrating strong robustness to guide perturbations.

\subsection{Cross-Dataset Generalization}\label{sec:cross_in_gen}
\vspace{-3pt}
The method comprises a segmentation model and a retrieval model. In previous experiments, both were trained and evaluated on the same dataset. To assess generalizability, we test whether J-RAS can improve performance when segmentation and retrieval models use different datasets. Specifically, segmentation is performed on M\&Ms, while retrieval uses guides from ACDC, evaluating the method’s ability to leverage out-of-distribution (OOD) guidance. As shown in \autoref{cross_in}, J-RAS improves segmentation performance in both in-distribution (retrieving and segmenting on M\&Ms) and cross-dataset (retrieving from ACDC while segmenting M\&Ms) settings. However, the improvement is more consistent and pronounced in the in-distribution case, while cross-dataset performance fluctuates and slightly degrades for higher Top-\textit{K} values due to domain mismatch, where the retrieved guides become less semantically aligned with the M\&Ms targets. This highlights that although J-RAS demonstrates generalization capability across datasets, it achieves the greatest benefit when the retrieval and segmentation domains are aligned.

\section{Conclusion, limitations, and future work}
\vspace{-7pt}
In this paper, we proposed Joint Retrieval-Augmented Segmentation \textbf{(J-RAS)}, a unified framework that redefines the relationship between segmentation and retrieval. Unlike prior approaches that treat retrieval as a static, similarity-based step, J-RAS establishes a mutual adaptation loop where both networks learn together, the segmentation model benefits from retrieved image–mask guidance, and the retrieval model is optimized through the segmentation loss to emphasize task-relevant features. By alternating contrastive and supervised learning, J-RAS transforms retrieval from a passive reference system into an active collaborator that provides context-aware guidance. Experiments on ACDC and M\&Ms using four backbones (SegFormer, U-Net, TransUNet, SAM) show consistent gains and strong cross-dataset generalization, confirming that J-RAS improves boundary precision and robustness across domains. Extensive ablations further validate its effectiveness under noisy guides and dynamic retrieval settings.

Despite its promising results, J-RAS has limitations. To accelerate training, retrieval updates are restricted to the query branch rather than propagating through the entire knowledge base, which may lead to minor mismatches for certain patients. Moreover, the approach relies on the diversity and representativeness of the retrieval set, which may constrain generalization to unseen modalities or underrepresented conditions. Future work will explore scalable end-to-end optimization, expanding and diversifying retrieval databases, and extending J-RAS to multi-modal and cross-domain scenarios. Such directions hold significant potential for translating retrieval-guided, contrastively optimized segmentation into robust, clinically deployable systems that learn and adapt much like human experts.

\bibliographystyle{elsarticle-num} 
\bibliography{ref}

\end{document}